\documentclass[journal,twoside,web]{IEEEtran}
\usepackage{cite}
\usepackage{amsmath,amssymb,amsfonts}
\usepackage{graphicx}
\usepackage{textcomp}
\usepackage{graphicx}
\usepackage{hyperref}
\usepackage{breakurl}
\usepackage{lineno}
\usepackage{amsmath}
\usepackage{subfig}
\usepackage{url}
\usepackage{float}
\usepackage{bm}
\usepackage{verbatim}
\usepackage{multirow}
\usepackage{amsmath,scalefnt}
\modulolinenumbers[5]
\usepackage{color}
\usepackage{multirow}
\usepackage[table]{xcolor}
\usepackage{adjustbox}
\usepackage{float}
\restylefloat{table}
\usepackage{placeins}
\graphicspath{{img/}}
\usepackage{soul}

\begin{document}
%
\title {Gigapixel Histopathological Image Analysis using Attention-based Neural Networks}

%
%
%

\author{Nadia~Brancati, Giuseppe~De~Pietro, Daniel~Riccio and Maria Frucci
\thanks{N.~Brancati, G.~De~Pietro, and M.~Frucci are with Institute for High Performance Computing and Networking of 
National Research Council of Italy (ICAR-CNR), Naples, Italy. e-mail: (nadia.brancati, giuseppe.depietro, maria.frucci)@cnr.it.}
\thanks{D.~Riccio is with Institute for High Performance Computing and Networking of
National Research Council of Italy (ICAR-CNR), Naples, Italy and with Universita' Federico II, Naples, Italy. e-mail: daniel.riccio@unina.it}
}

\maketitle

\begin{abstract}
Although CNNs are widely considered as the state-of-the-art models in various applications of image analysis, one of the main challenges still open is the training of a CNN on high resolution images. Different strategies have been proposed involving either a rescaling of the image or an individual processing of parts of the image. Such strategies cannot be applied to images, such as gigapixel histopathological images, for which a high reduction in resolution inherently effects a loss of discriminative information, and in respect of which the analysis of single parts of the image suffers from a lack of global information or implies a high workload in terms of annotating the training images in such a way as to select significant parts. We propose a method for the analysis of gigapixel histopathological images solely by using weak image-level labels. In particular, two analysis tasks are taken into account: a binary classification and a prediction of the tumor proliferation score. Our method is based on a CNN structure consisting of a compressing path and a learning path. In the compressing path, the gigapixel image is packed into a grid-based feature map by using a residual network devoted to the feature extraction of each patch into which the image has been divided. In the learning path, attention modules are applied to the grid-based feature map, taking into account spatial correlations of neighboring patch features to find regions of interest, which are then used for the final whole slide analysis. Our method integrates both global and local information, is flexible with regard to the size of the input images and only requires weak image-level labels. Comparisons with different methods of the state-of-the-art on two well known datasets, Camelyon16 and TUPAC16, have been made to confirm the validity of the proposed model.  

\end{abstract}


%
\IEEEpeerreviewmaketitle

\section{Introduction}
\label{sec:intro}

Histopathology has played a vital role in cancer diagnosis and prognostication for over a century. Nowadays, slide-scanning microscopes provide digital whole slide images (WSIs) of the digitization of patient tissue samples, which allow pathologists to collaborate rapidly and remotely for diagnostic, teaching and research purposes. With the increasing capability of a routine and rapid digitization, the application of Artificial Intelligence techniques to WSIs can provide an automatic image analysis of the tissue morphology with the potential to assist pathologists to be more productive, objective and consistent in diagnosis. In particular, deep networks have produced groundbreaking results in many tasks related to digital pathology, for example segmentation, nuclei detection and classification. Moreover, differently from traditional machine learning approaches that require the design of hand-crafted features by domain experts, as an end-to-end model deep networks are able to learn the best features to provide the desired results by directly training on input raw images. However, the automatic processing of histopathological images still poses extremely complex challenges. Crucial problems arise from the need to process gigapixel images (e.g. 2GB or more) with a high spatial resolution, a limited signal to noise ratio and different levels of magnification. Due to the large size of the WSIs, the direct application of conventional deep learning techniques is precluded by the current state of the available hardware. On the other hand, the challenge of gigapixel image analysis cannot be addressed by means of an extensive image down-sampling, as this would cause a dramatic loss of discriminative features. 
The first attempts to apply deep learning models to WSI classification were based on the assumption that the label at image-level could be inferred by combining the labels obtained at patch-level independently. According to this simplification, the whole WSI image is partitioned into a number of patches small enough to be independently processed by a deep network.  A target class for the entire WSI is then inferred by combining the decisions obtained for the single patches. Such a deep learning strategy requires that the labels provided at patch-level are properly trained. Moreover, a lot of patch labels are needed to produce the generic CNN features able to capture the heterogeneity of some cancer subtypes. However, most datasets commonly used for training deep neural networks only provide ground truth labels for the WSIs, since patch-level annotation in gigapixel images is a tedious, time consuming and error-prone process. 
In order to cope with the limited availability of labeled data, some works in literature assume that the image label could be assigned to all patches extracted from the image~\cite{spanhol2016breast, araujo2017classification,vang2018deep,gecer2018detection,duran2020}. This assumption neglects the fact that tumors may consist of a mixture of structures and texture properties, with the results that the patch-level labels are not necessarily consistent with the image-level label (e.g. many patches of a WSI with a cancer label may contain only healthy tissue). Noisy ground truth labels mislead deep learning models during the training process, so jeopardizing the performance of classification approaches implementing simple decisions based on methods involving the fusion of single patch predictions (e.g. voting and max-pooling)~\cite{kong2009computer}. In order to mitigate the effect of noisy ground truth labels on classification performance, decision fusion models specifically trained for the smart aggregate patch-level predictions given by patch-level CNNs have been proposed~\cite{hou2016patch, vang2018deep}.
In~\cite{hou2016patch} all the patch-level predictions are used to train a multi-class logistic regression that is shown to outperform max-pooling in predicting the image-level label, while in~\cite{vang2018deep} logistic regression is combined with a majority voting and gradient boosting machine to build an even more accurate ensemble fusion framework.
Alternatively, multiple instance learning techniques have been applied to select the patches to be analyzed for the prediction of the whole slide labels~\cite{cruz2018high,srinivas2014simultaneous,vu2015histopathological,das2020}, as the mere presence of malignant patches is considered sufficient to make a prediction at the image level. In~\cite{cruz2018high}, a CNN tile classifier is adopted to implement an adaptive sampling method for the precise detection of invasive breast cancer on WSIs. Srinivas et al.~\cite{srinivas2014simultaneous} have implemented a blob detection approach to extract only cellular patches, which are independently encoded by a sparsity model. Vu et al.~\cite{vu2015histopathological} further improved this model by proposing a class-specific dictionary learning method for the sparsity encoding. 
All these approaches can only take into account patterns present within individual patches, neglecting the potential relationships between them, a knowledge of which is necessary to obtain the global features at the image level. Considering that small patches share spatial correlations with their neighboring patches, ignoring these relationships makes the prediction of the CNN an isolated result. Huang et al. cope with
this problem by designing a Deep Spatial Fusion Network~\cite{huang2018improving}. This model implements a modified Residual Neural Network (ResNet) to compute patch probabilities that are arranged in a grid according to the patch-wise spatial order. A deep multi-layer perceptron (MLP) is trained to predict the image-level label from the spatial distribution of probability maps. Although patch decision fusion generally increases the image classification performance, it is well known that most of the discriminative information is lost at decision level. In order to overcome this limitation, recent research suggests aggregating patch data at feature level rather than at decision level. 
In~\cite{wang2018weakly}, a patch-based CNN is used to find discriminative regions for which context-aware features are selected by imposing spatial constraints, and then aggregated to feed a Random Forest classifier. Mercan et al.~\cite{mercan2019patch} also train a CNN to extract patch features, but they consider both feature vectors and class probabilities to construct an ROI representation. The ROI vector is generated by concatenating the features of the patches included, weighted by the corresponding class probabilities, and a MLP classifier is trained to predict the ROI decisions. Courtiol et al.~\cite{courtiol2018classification} propose a context-aware classification from tile instances by adding an additional set of fully-connected layers to the DCNN proposed in~\cite{he2016deep}. Being part of the proposed end-to-end model, these layers can still be trained on massive WSIs thanks to a random tile sampling scheme and to a strict set of regularizations.
These approaches rely heavily on the performance of the patch selection strategy, which may fail to identify malignant samples when they are in a limited number in the WSI. 
Aiming at preserving the spatial correlations in the whole WSI, Tellez et al.~\cite{tellez2019neural} have proposed a Neural Image Compression technique that maps images from a low-level pixel space to a higher-level latent space using neural networks. Features are extracted patch-wise by an encoder and rearranged to form a compressed image that saves most of the discriminative information and can be used to train a CNN to classify the entire WSI. Along the same lines, Tomita et al.~\cite{tomita2019attention} designed a grid-based convolutional attention-based mechanism that is trainable end-to-end. The attention module implements a 3-dimensional convolution operation and it is able to process input images of any rectangular shape. The attention modules are combined with the extracted features to predict the label at the image level.  
The works in~\cite{tellez2019neural} and~\cite{tomita2019attention} demonstrate that a more compact representation of the WSI retains a sufficient quantity of features to predict the image label with a high accuracy. In~\cite{tomita2019attention}, the deep network adopted to extract features from the patches is part of the end-to-end model, so increasing the number of parameters to be trained. In order to curtail the computational burden of the training process, only the last levels of the feature extraction network are trained with a batch of very limited dimensions (two images). Moreover, the input WSI is split into smaller sub-images containing tissue regions labeled coherently with the image-label and resulting to be ease to manage thank to their smaller dimensions. However, in~\cite{tellez2019neural} the authors show that it is possible to exclude the feature extraction network from the whole model by using a pre-trained network to extract features from the patches and training only the CNN for the classification of the neural compressed image. Since the adopted CNN is not able to process inputs of different size, compressed images of fixed size are generated by extracting several independent crops of the same size from the input WSI.
Starting from the interesting results obtained in~\cite{tellez2019neural} and~\cite{tomita2019attention}, in this work we move a step forward, as we have not only taken the feature extraction module out of the end-to-end model, but we have also taken advantage of this to design a more complex classification network able to process a WSI of any size. This last represents a quite desirable property that saves our model from applying any resizing process to the input and so retaining the whole informative content of the original WSI. Indeed, we have used a simple pre-trained network to extract the patch features, while introducing two levels of attention in the classification network to achieve a better selection of the features extracted to predict the labels at the image-level. Experiments have been conducted on two
different tasks, namely breast cancer binary classification and breast tumor proliferation speed prediction. Comparisons with state-of-the-art methods confirm that the combination of two sets of attention maps with the extracted features produces an increase in the image classification accuracy. 
The rest of the paper is organized as follows: in Section \ref{sec:method} a description of the proposed approach is given; the experimental setup, comparative strategies and results for the two different tasks are reported in Section \ref{sec:experiments}; discussions about the obtained results are given in Section \ref{sec:discussion}. Finally, in Section \ref{sec:conclusion} the conclusions are drawn.

\section{Method}
\label{sec:method}
\begin{figure*}

      \centering
      \includegraphics[width=400pt]{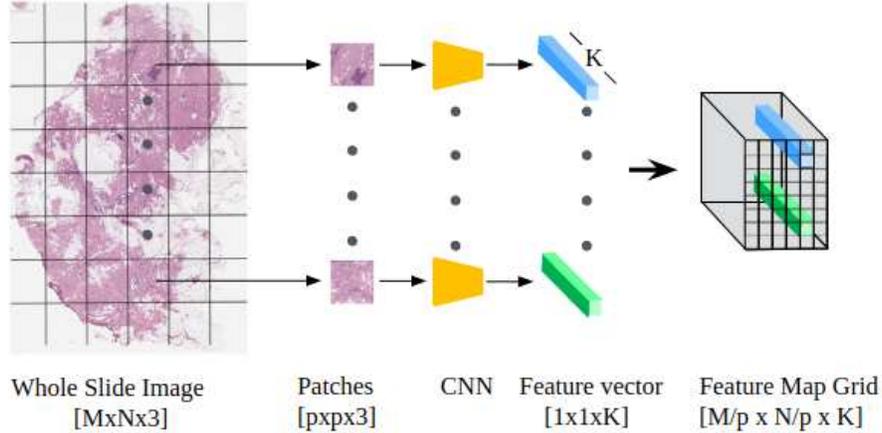}
      \caption{Grid-based Feature Extraction. A WSI is divided into a set of patches and each of them is mapped to a feature vector using a CNN. The set of feature vectors is rearranged in a grid-based feature map according to the original spatial arrangement of the patches.}
      \label{figure:method1}
\end{figure*}

The overall framework is functionally divided into two main stages, namely Grid-based Feature Extraction (GFE) and Attention-based Classifier (AC) (see Figures~\ref{figure:method1} and~\ref{figure:method2}). The former is devoted to mapping the WSI in a new compressed and dense feature space, while the latter applies an attention-based mechanism to weight the extracted features, which are then fed to the image-level classifier. Inspired by the attention-based method presented in~\cite{tomita2019attention}, the feature extraction network and the attention-based mechanism are kept separate and not linked in an end-to-end fashion. In other words, our GFE applies a CNN to extract patch-wise features and aggregate feature vectors in a compact grid representation according to the spatial location of the corresponding patches in the WSI. The AC implements both the min- and max-attention mechanisms separately on the input grid-based feature map and produces two different sets of attention maps. The attention maps are used to drive the classification process, as they lead the classifier to focus on features that are considered more expressive for the class learned. In our framework, the AC is the only part involved in the training process and its outcome is an image-level label.

\subsection{Grid-based Feature Extraction}
The input of the system is a WSI gigapixel image of a rectangular shape. The purpose of the GFE is to produce a more compact representation that could be managed by the following processing pipeline in its entirety. To achieve this aim, the GFE maps the original pixel based representation of the image into a low dimensional feature space by preserving local spatial relationships. The WSI is partitioned into a set of non overlapping patches that are mapped into a feature vector by applying a CNN. The feature vectors are rearranged according to a grid-based feature map, preserving the spatial proximity information which the patches present in the WSI input image. 
In more detail, let $W\in R^{M\times N\times3}$ denote an input image with $M$ rows, $N$ columns, and three color channels (RGB). The GFE extracts a set $X =\{x_{i,j}\}$ of non-overlapping patches by sampling $W$ along the $i-th$ row and $j-th$ column according to a uniform grid of size $p\times p\times3$. Each patch $x_{i,j}$ is independently fed to a CNN and the out-coming values of the global average pooling layer of the network are taken as a $1\times 1\times K$ patch-wise feature vector. The GFE stacks the features vectors into a three-dimensional compressed representation according to a grid $G\in R^{M'\times N'\times K}$ with $M'=M/p$ and $N'=N/p$, so that the  element $g_{i,j}$ is the feature vector of $x_{i,j}$ and adjacent elements of $g_{i,j}$ in $G$ correspond to adjacent patches of $x_{i,j}$ in $W$. Figure~\ref{figure:method1} shows a schematic representation of the GFE module. 
The CNN plays a crucial role in the system, as it provides the features which the rest of processing pipeline will work on. Thus, different CNNs have been investigated and ResNet34 has proved to be the best performing one.

\subsection{Attention-based Classifier}
The Attention-based Classifier is the combination of a 3D Convolutional Layer (3DCL) with two different attention modules. The 3DCL convolves the compressed image with a set of filters along its depth $K$. The aim of this module is twofold, as it merges information provided by the features of neighboring patches and gathers it to a shallower representation. This new compact volume is fed independently to two attention modules, each of which producing a set of two-dimensional attention maps. The spatial resolution of each attention map is equivalent to the first two dimensions of $G$, so that its elements can be seen as weighting the contribution provided by the single patches in determining the final label assigned to the whole image by the classifier. In other words, the attention maps provide information on image regions that are considered either significant or irrelevant for the classifier to take the final decision.  
In more detail, the 3DCL consists of a kernel filter with a size $n\times n\times K$ that is convolved with $G$ along its third dimension to produce a new grid-based feature map $G'$ with size $M'\times N'\times H$, where $H<K$. 
Max-pooling and min-pooling are then applied on the output of the 3DCL independently, so obtaining two different sets of feature maps, $Mp$ and $mp$ respectively, each with a size $M’\times N’\times H$. Without any loss of generality, we focus on $Mp$ to detail the attention mechanism, as it is equivalent for $mp$. 
The attention maps $AMp$ ($Amp$) are computed element-wise from $Mp$ ($mp$) by applying the softmax operator $\sigma$:
\begin{equation}
AMp_{i,j,h}=\sigma(Mp_{i,j,h})= \frac{e^{Mp_{i,j,h}}} {\sum_{m=1}^{M'} \sum_{n=1}^{N'} e^{Mp_{m,n,h}}}   
\end{equation}

\noindent with $i=1,…,M’$,  $j=1,…,N’$, $h=1,...,H$ and $\sum_{i=1}^{M'} \sum_{j=1}^{N'} \sigma(Mp_{i,j,h})=1$.\\

\begin{figure*}

     \centering
     \includegraphics[width=400pt]{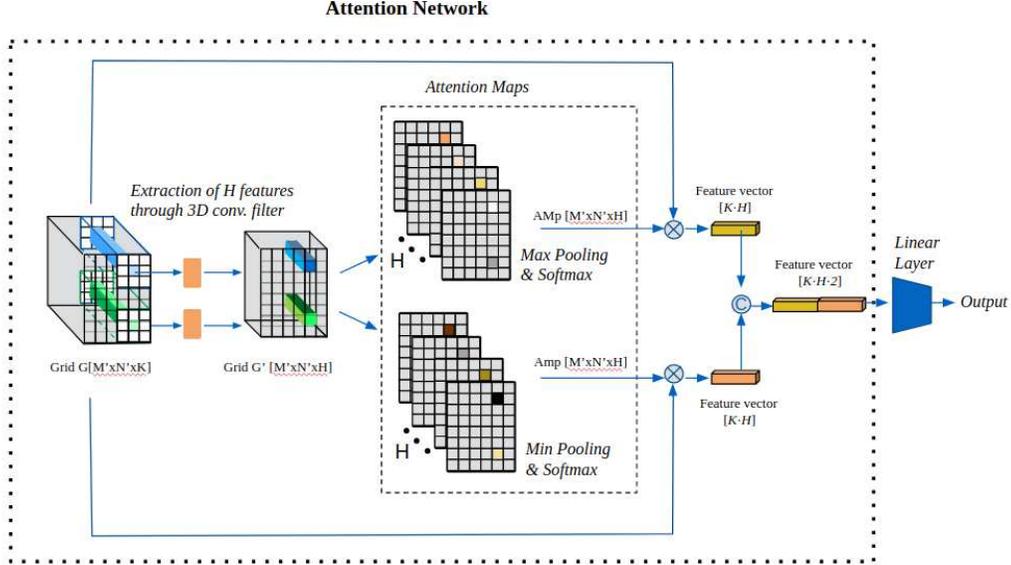}
     \caption{Our attention-based network. First, 3D convolutional filters of size $n\times n\times K$ are applied to the grid-based feature map $G$ with size $M’\times N’\times K$ to generate a new grid-based feature map $G’$ with size $M’\times N’\times K$, with $H<K$. Secondly, two different mechanisms of pooling, max and min pooling, are applied on $G’$ independently to produce two sets of attention maps $AMp$ and $Amp$. Thirdly, the sets of the attention maps operate as the weights for affine combinations of the initial grid-based feature map $G$ producing two feature vectors with $K\cdot H$ size that are concatenated to feed a linear layer for the classification/regression tasks.}
      \label{figure:method2}
\end{figure*}

The attention maps are then used to weight features in $G$, so to compute a feature vector of size $H\times K$ according to the following equation:
\begin{equation}
v_{k,h}=\sum_{m=1}^{M'} \sum_{n=1}^{N'} AMp_{m,n,h} \cdot G_{m,n,k} 
\end{equation}
with $k=1,...,K$ and $h=1,...,H$. A non-linear activation function is then applied to $v$. Since two different sets of attention maps are considered separately, two feature vectors are obtained for $G$ at the end of this process. Thus, the two feature vectors are concatenated to form a feature vector representing the whole image and this last is fed to a linear layer that produces an image-level label. In Figure~\ref{figure:method2}, a scheme of the proposed attention network is shown.

\section{Experiments and results}
\label{sec:experiments}
A series of experiments has been performed to assess the performance of the proposed method on two publicly available histopathology image datasets, namely Camelyon16~\cite{bejnordi2017diagnostic} and Tumor Proliferation Assessment (TUPAC16)~\cite{veta2019predicting}. In particular, different strategies have been considered for the training of the CNN adopted in the GFE module, aimed at assessing the potential contribution of fine-tuning. Moreover, three attention mechanisms have been tested for the Attention Classifier, namely min, max, and average. Experiments have been conducted to validate the accuracy of the classifiers, both when they are considered alone and when two of them are combined together. The performance of the proposed system has been compared with that provided by state-of-the-art techniques on the same tasks with respect to the same testing protocols.

\subsection{Histopathology Datasets and Data Preparation}

The Camelyon16 dataset consists of $400$ hematoxylin and eosin (H\&E) WSIs of sentinel lymph nodes of breast cancer obtained from two independent datasets that have been collected in Radboud University Medical Center (Nijmegen, the Netherlands) and in the University Medical Center Utrecht (Utrecht, the Netherlands). For our experiments, we adopted the original splitting of the dataset, i.e. $270$ WSIs for the training and $130$ WSIs for the testing. We evaluated our model with respect to the binary classification problem normal/malignant, only by using image-level labels.

The TUPAC16 dataset is composed of $492$ breast cancer cases from The Cancer Genome Atlas that are generally used to train and evaluate a model. Additionally, $321$ test WSIs with no public ground truth are also available for an independent testing that is directly performed by the chairs of the TUPAC16 challenge on request. The experiments which we performed on TUPAC16 relate to the prediction of the proliferation score based on molecular data. For this task, also, we adopted the original splitting of the dataset for the training and testing and we used only image-level labels. 

The image-level label associated to the WSIs is the presence/absence of tumor metastasis for Camelyon16 and the degree of tumor proliferation speed based on molecular data for TUPAC16. For both datasets, the WSIs were processed at a $2$ $\mu m/pixel$ resolution. All the WSIs underwent a preprocessing step aimed at detecting the white background on the slide and at identifying as the image to be treated the smallest box including all tissue regions. An augmentation of each training dataset was performed by considering a set of $9$ transformations for each WSI (i.e. one horizontal flip, one vertical flip, three clockwise rotations of $90^{\circ}$, two horizontal translations and two vertical translations). In order to further extend the training set, the WSIs of the training datasets at a $4$ $\mu m/pixel$ resolution were also included.
The BreakHis dataset \cite{spanhol2015dataset} represents a further testbed to validate classification approaches of breast cancer histopathological images. However, we have not considered this dataset in our experiments, as our method works on whole slide images with very high spatial resolution, while BreakHis provides images of $700\times460$ pixels. Since a number of $224\times224$ non overlapping patches are needed to arrange our grid, it would make little sense to test it on BreakHis.

\subsection{Network architectures and training protocols}
As the feature extractor in the GFE module, we selected the residual neural network ResNet architecture~\cite{he2016deep}, as it shows lower complexity than other popular CNNs, such as VGG-Net or Google-Net, while still obtaining a good trade-off between performance and GPU memory usage. More specifically, ResNet-34 was considered in this study. The classic ResNet-34 model is pre-trained on a very large natural image dataset, namely ImageNet~\cite{imagenet_cvpr09}. Additionally, we also investigated the application of a fine-tuning strategy on the pre-trained ResNet-34 on the Breast Cancer Histology Images (BACH) dataset~\cite{aresta2019bach} as performed in~\cite{brancati2018multi}. As shown in Section 3.4, the fine-tuned ResNet-34 provided better features when compressing the WSI for the classification system. For the sake of simplicity, in the following sections the Attention-based Classifier will be denoted as PT-R34-AC or FT-R34-AC, depending on whether it is based on features coming from the pre-trained ResNet-34 or the fine-tuned ResNet-34 network. The system inputs a WSI that is split into a set of non overlapping patches with a size equal to $224\times224$. The CNN maps each patch into a vector of $K=512$ features that will be part of the compressed representation of the image.

The attention-based network was trained for $35$ epochs with a batch size equal to $64$ for both datasets. Different loss functions and optimizers were adopted, depending on whether the task to be performed was to predict the presence of metastasis or to estimate the tumor proliferation speed at image level. In particular, for the Camelyon16 dataset, the Cross Entropy Loss was considered for the backpropagation and an SGD optimizer was adopted with an initial learning rate equal to $0.001$ with a decay of $0.1$ after $20$ epochs. For the TUPAC16 dataset, logistic regression was applied to predict the proliferation score at WSI level, so the Mean Square Error Loss was considered for the backpropagation, while an Adam optimizer with an initial learning rate equal to $0.0001$ was used for the optimization. 

\subsection{Performance comparisons and visualization of the discriminative features on the Attention Maps}
We evaluated the performance of the proposed model with respect to two different tasks: 1) predicting the presence of metastasis and 2) predicting the tumor proliferation speed at image level. Moreover, we compared the performance of the PT-R34-AC and FT-R34-AC models with state-of-the-art approaches that have been designed for the above-mentioned classification tasks. In particular, we compared these with the model proposed in~\cite{tellez2019neural}, according to the same experimental protocol, namely a 4-fold cross validation and two random weight initializations. In order to obtain results which are comparable, the AUC metric and Spearman correlation coefficient were computed for the first and second classification tasks, respectively. Comparisons with the model proposed in~\cite{tomita2019attention} are also provided. For this evaluation, we have adopted our own implementation of the approach~\cite{tomita2019attention} because no code is available for the comparisons and the performance provided in ~\cite{tomita2019attention}  have been obtained on a private dataset. The method has been implemented in Pytorch and was run on an NVIDIA GeForce GTX 1080TI 11 GB RAM.

As already mentioned, all the experiments for a given classification task involved the splitting of the dataset into four equal-sized partitions that are differently selected into four rounds of cross-validation using three partitions for the training and one for the validation. 
As regards Camelyon16, the testing was performed on the original test set. On the contrary, for TUPAC16 we report the results obtained on the validation set, due to difficulty in obtaining results on testing set from the organizers of the challenge. The mean and standard deviation of the selected evaluation metrics using two random weight initializations were computed.
The results in Table~\ref{table:table1} show the AUC values for the experiments related to the task 1. It is worth noting that the FT-R34-AC model achieved a performance of $0.711$ AUC, so outperforming the method in~\cite{tellez2019neural} and in~\cite{tomita2019attention}. The PT-R34-AC model provided less impressive results with respect to~\cite{tellez2019neural}, but still outperforms~\cite{tomita2019attention}. In order to assess the statistical significance of experiments, we performed the Wilcoxon test~\cite{wilcoxon1992individual} between FT-R34-AC and PT-R34-AC models. Analyzingthe results at a $0.05$ significance level, the computed p-value equal to $0.01$  attests a statistical significance difference between the two classifiers. Similarly, we have performed the Wilcoxon test between FT-R34-AC (PT-R34-AC) and the method in~\cite{tomita2019attention} obtaining a p-value equal to $0.01$ ($0.02$) that attests the statistical significance difference between the two classifiers.

\begin{figure*}
\centering
   \captionsetup[subfigure]{labelformat=empty}
      {\includegraphics[width=400pt]{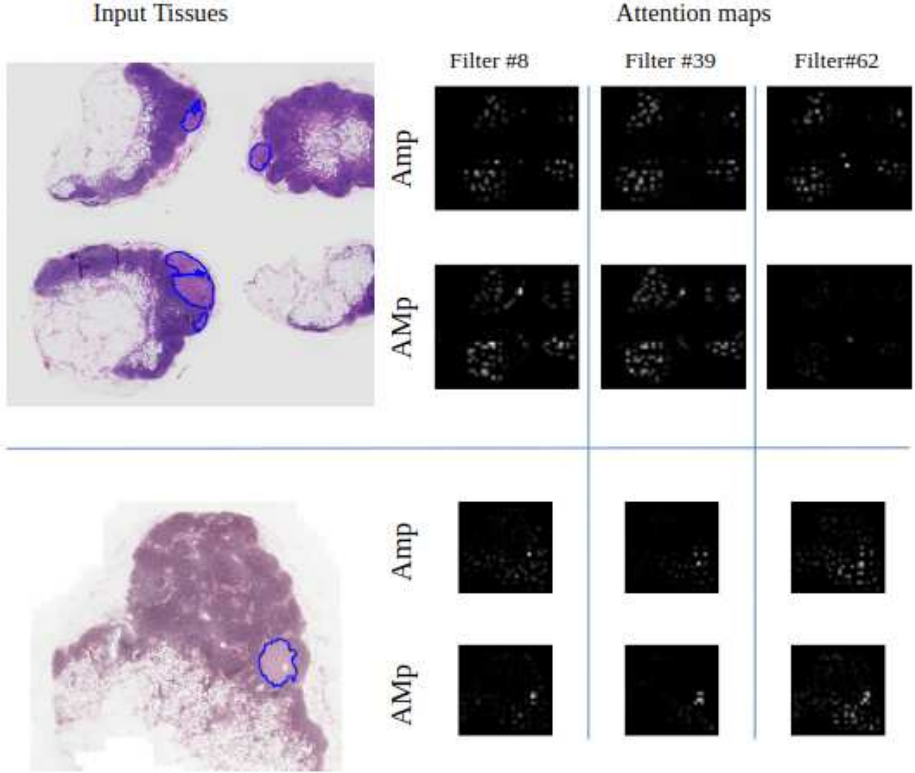}}

      \caption{Examples of some attention maps for Camelyon 16, generated by different sets of attention modules which highlight specific features of the tumor class. The first column shows WSIs from the test set for which the tumor location is underlined in blue. The second to fourth columns show the selected attention modules for the input images from different set of attention maps (i.e., the first and third rows for the max-pooling-based attention maps; the second and fourth rows for the max-pooling-based attention maps). The highest attention weight is denoted in white color and the lowest in black.}
      \label{figure:AM1}
\end{figure*}

\begin{figure*}
\centering
   \captionsetup[subfigure]{labelformat=empty}
      
      {\includegraphics[width=400pt]{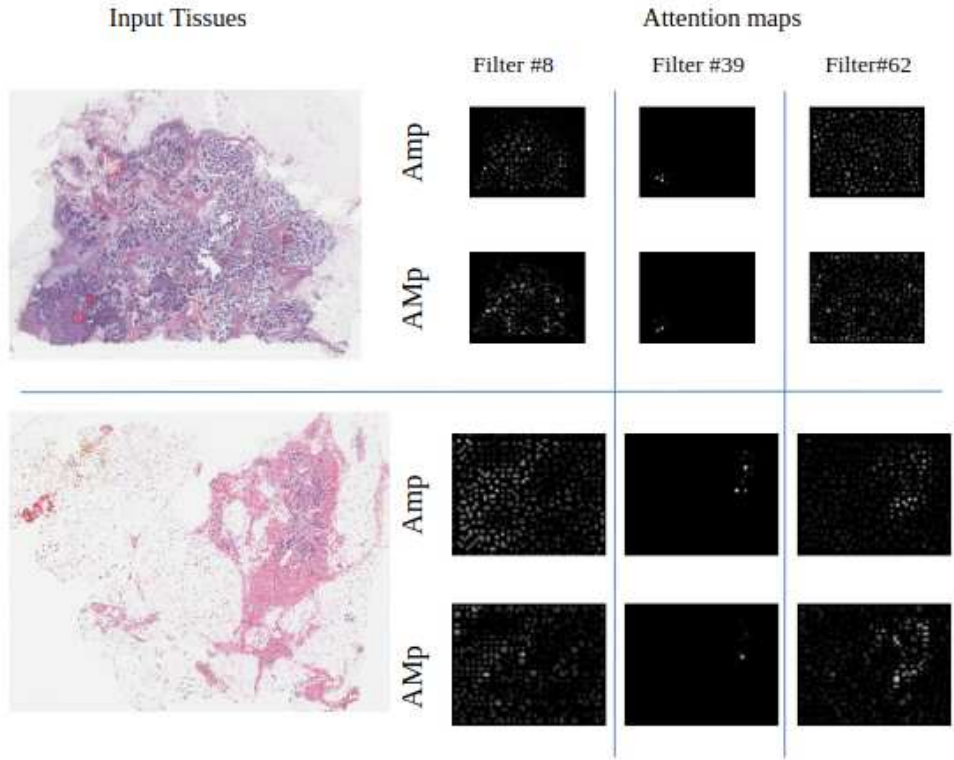}}

      \caption{Examples of some attention maps for TUPAC16}
      \label{figure:AM2}
\end{figure*}

In Figure~\ref{figure:AM1} (\ref{figure:AM2}), we report some examples of the $64$ attention filters for each of the two sets of attention maps that were generated by FT-R34-AC (PT-R34-AC) for two different WSIs with tumors, for dataset Camelyon 16 (TUPAC16). The filters of the set of the max pooling-based attention maps ($AMp$) highlight specific features on critical regions, while the filters of the set of the min pooling-based attention maps ($Amp$) compute a complementary attention inducing a more effective feature learning.
Indeed, it can be noted from Figures~\ref{figure:AM1} and ~\ref{figure:AM2} that $Amp$ provides a lower but more diffused response, so leading the model to consider a higher number of patches but with a lower confidence degree.  

\begin{table} [!hb]
\centering
\resizebox{0.4\textwidth}{!}{
	\begin{tabular}{ | c | c |} 
	
	\hline   \textbf{Method} & \textbf{AUC}  \\ 
	\hline Method in \cite{tomita2019attention} &  $0.553 (0.040)$ \\
	\hline Method in \cite{tellez2019neural} &  $0.704 (0.030)$ \\
	\hline PT-R34-AC & $0.590 (0.009)$ \\ 
	\hline FT-R34-AC & \boldsymbol{${0.711 (0.001)}$} \\ 
	\hline
	\end{tabular}}
	\caption{Predicting the presence of metastasis at WSI level. Mean and standard deviation of the AUC values computed using two random weight initializations.}
	\label{table:table1}
\end{table}

The results in Table~\ref{table:table2} show the Spearman correlation scores for the experiments performed for task 2. 
Both FT-R34-AC and PT-R34-AC performed better than~\cite{tellez2019neural}, while their Spearman score is lower than that of the method in~\cite{tomita2019attention}. However, the Wilcoxon testsuggests that for task 2 there is no statistical significance difference between our models and Tomita's method, as the computed p-value is greater than $0.05$ at a $0.05$ significance level. Furthermore, Tomita's method required a large amount of resources in terms of time and memory compared with a little improvement in terms of Spearman score. This is consistent with what the authors in~\cite{tomita2019attention} point out by saying that the batch size is limited to 2 images since the feature extractor is included in the entire training loop.
As regards which feature extractor model is adopted (pre-trained or fine-tuned ResNet34), we have performed the Wilcoxon test between FT-R34-AC and PT-R34-AC models also for this experiment. Analyzing the results at a $0.05$ significance level, the computed p-value greater than $0.05$  points out there is no statistical significance difference between the two classifiers. However, both models perform coherently with respect to~\cite{tellez2019neural} and~\cite{tomita2019attention} in terms of Spearman correlation score.

\begin{table} [!hb]
\centering
\resizebox{0.4\textwidth}{!}{
	\begin{tabular}{ | c | c |} 
\hline   \textbf{Method} & \textbf{Spear. score}  \\ 
    \hline Method in \cite{tomita2019attention} &  \boldsymbol$0.630 (0.020)$\\
	\hline Method in \cite{tellez2019neural} &  $0.522 (0.001)$ \\
	\hline PT-R34-AC & $0.619 (0.009)$\\ 
	\hline FT-R34-AC & $0.596 (0.009)$ \\ 
	\hline
	\end{tabular}}
	\caption{Predicting the presence of metastasis at the WSI level. The mean and standard deviation of the AUC values were computed using two random weight initializations.}
	\label{table:table2}
\end{table}

\section{Discussion}
\label{sec:discussion}

\begin{figure*}

\centering
\captionsetup[subfigure]{labelformat=empty}
     {\includegraphics[width=190pt]{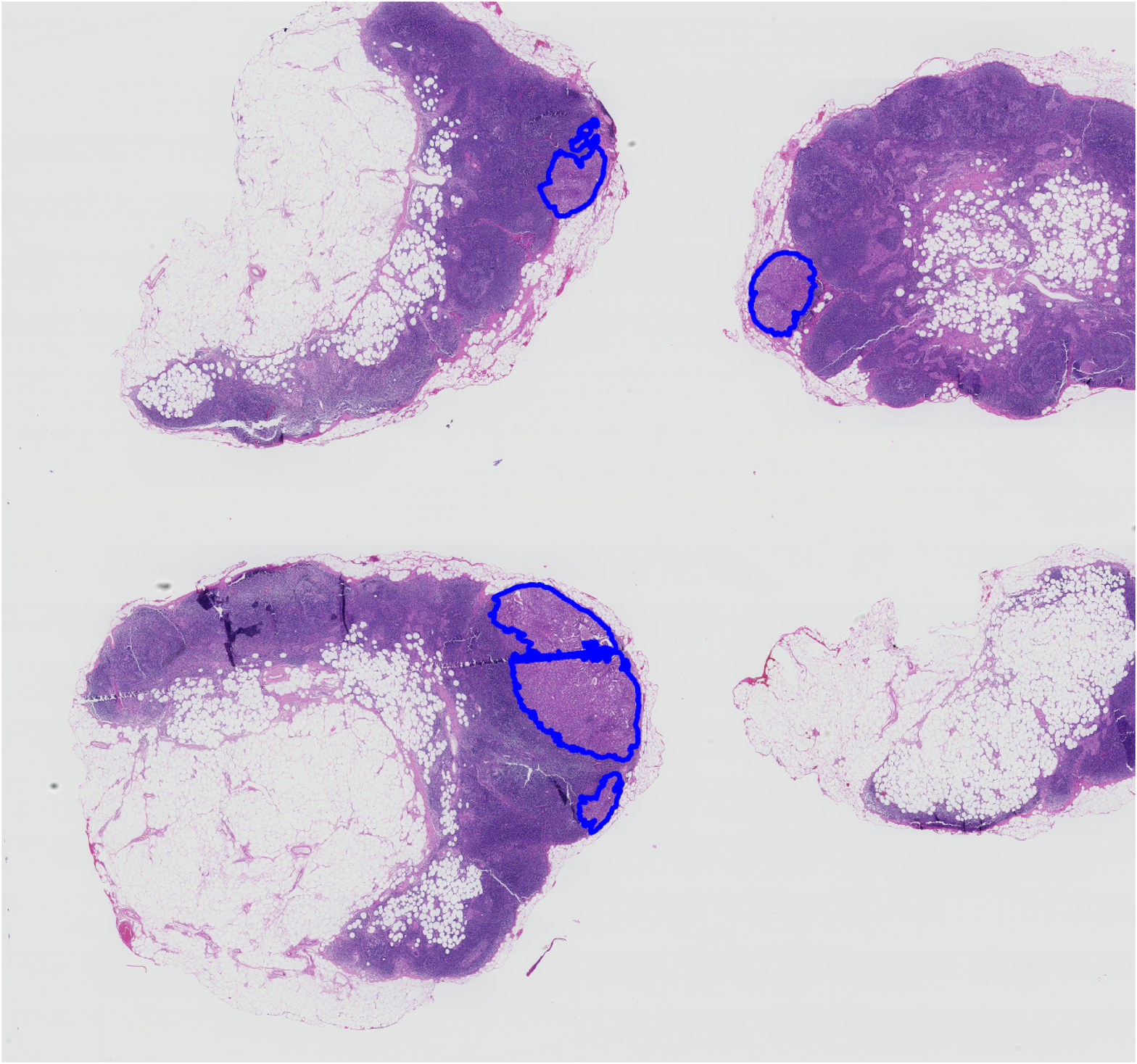}}
      {\includegraphics[width=182pt]{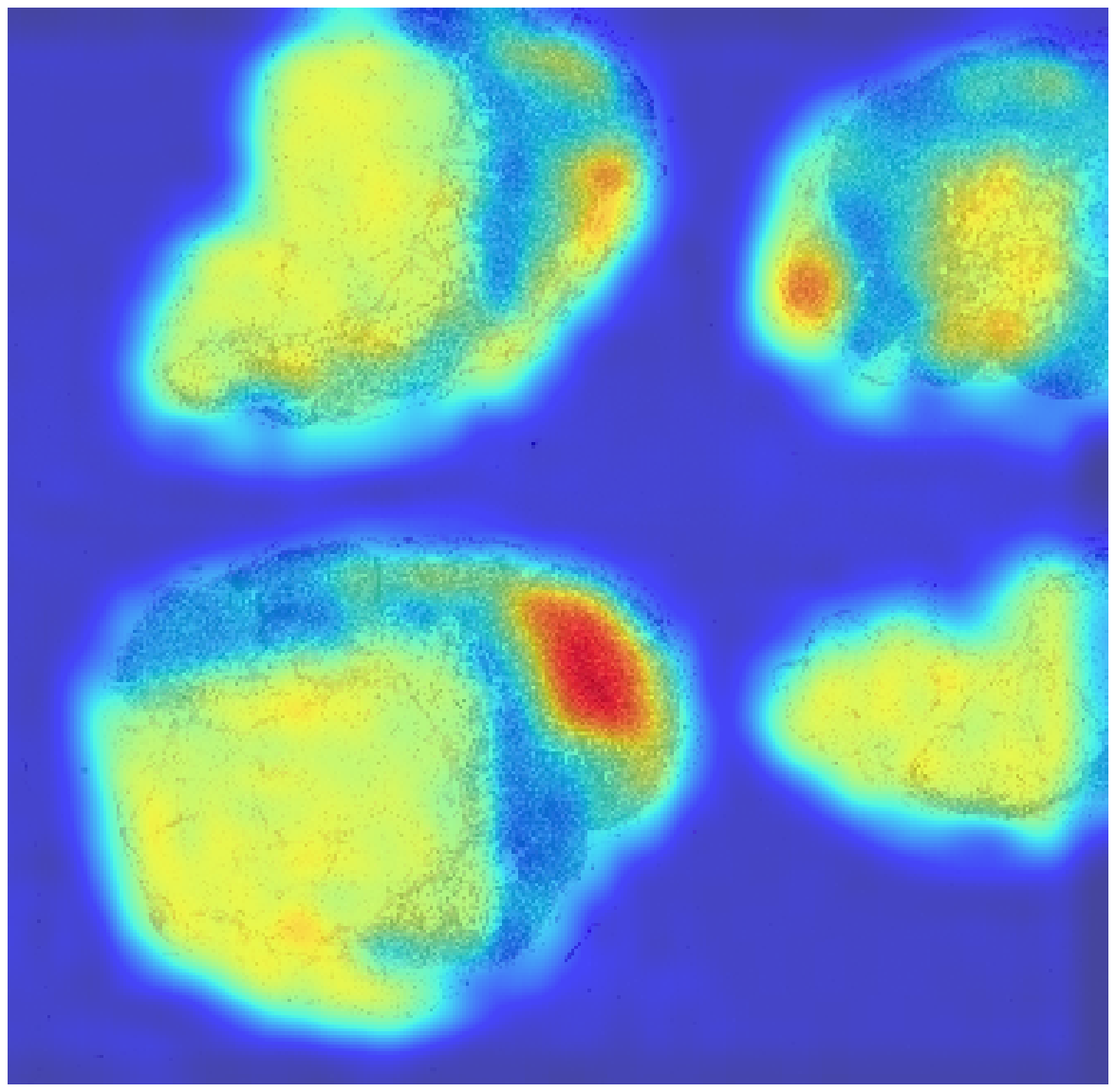}}\\
      \vspace{2cm}

      {\includegraphics[width=174pt]{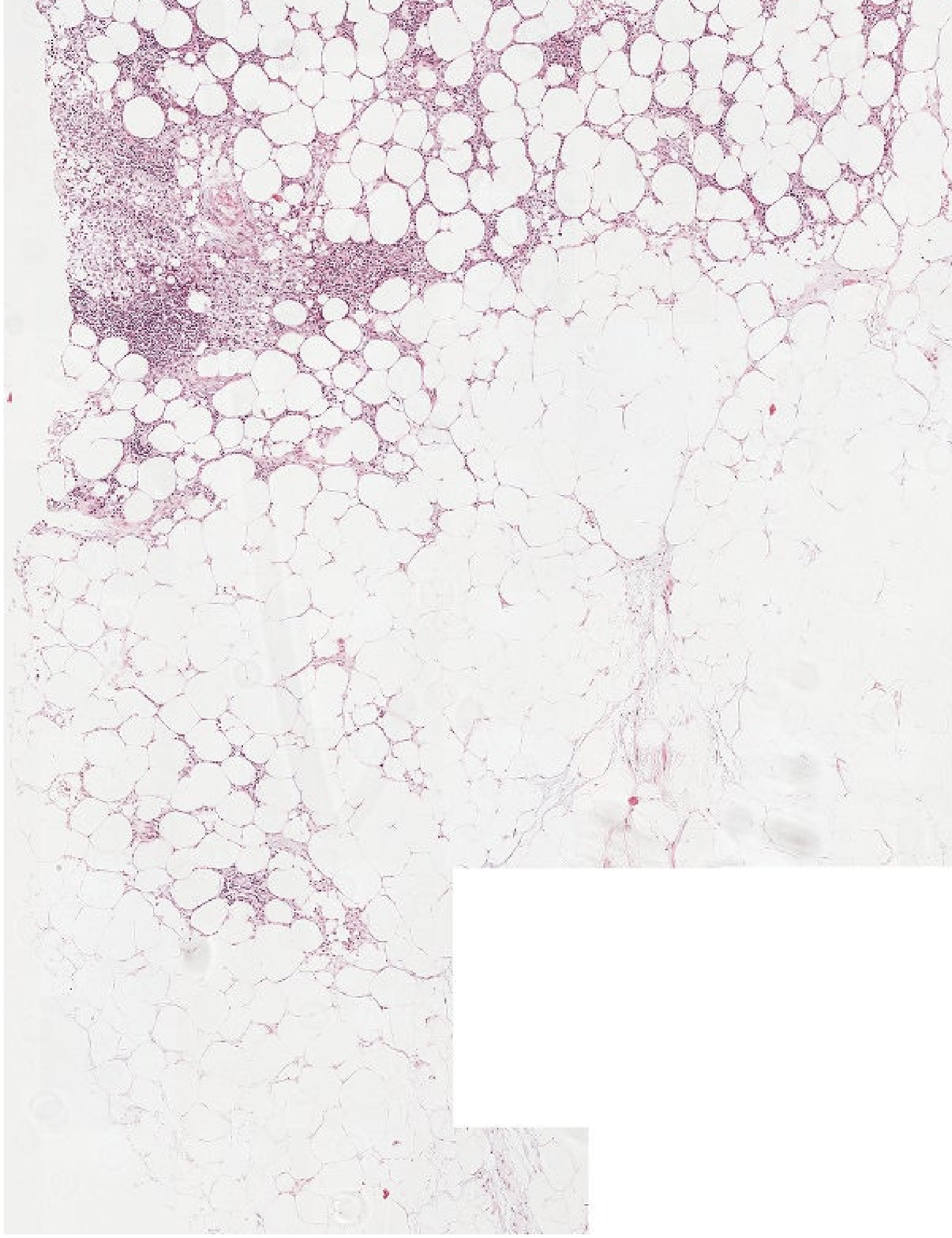}}
      {\includegraphics[width=170pt]{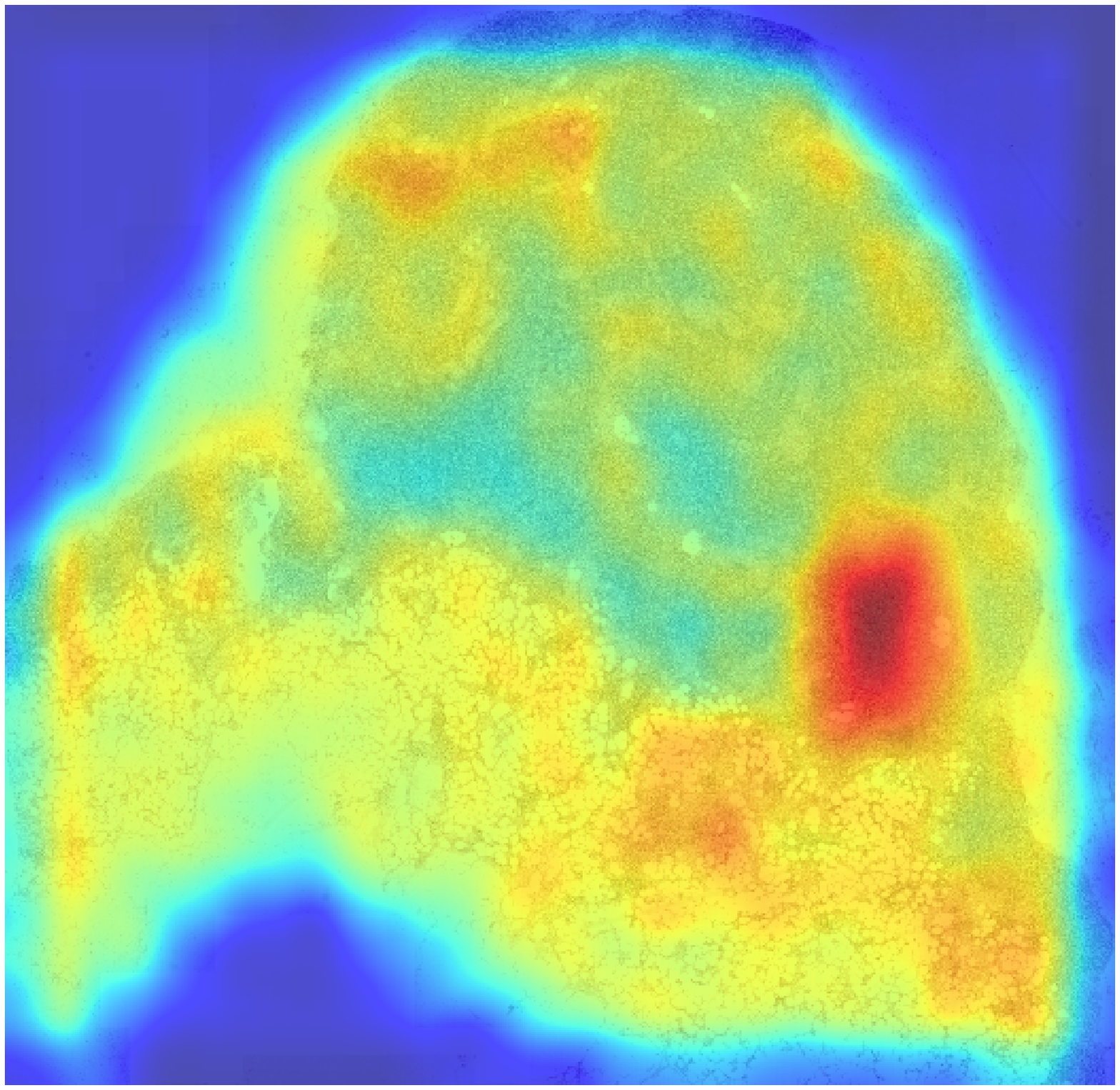}}\\
      
      \caption{Grad-CAM visualization applied to a sample WSIs from Camelyon 16. Dark blue coloring indicates a low saliency, whereas red indicates a high saliency.}
      \label{figure:Saliency1}
\end{figure*}

\begin{figure*}

\centering
\captionsetup[subfigure]{labelformat=empty}
     {\includegraphics[width=190pt]{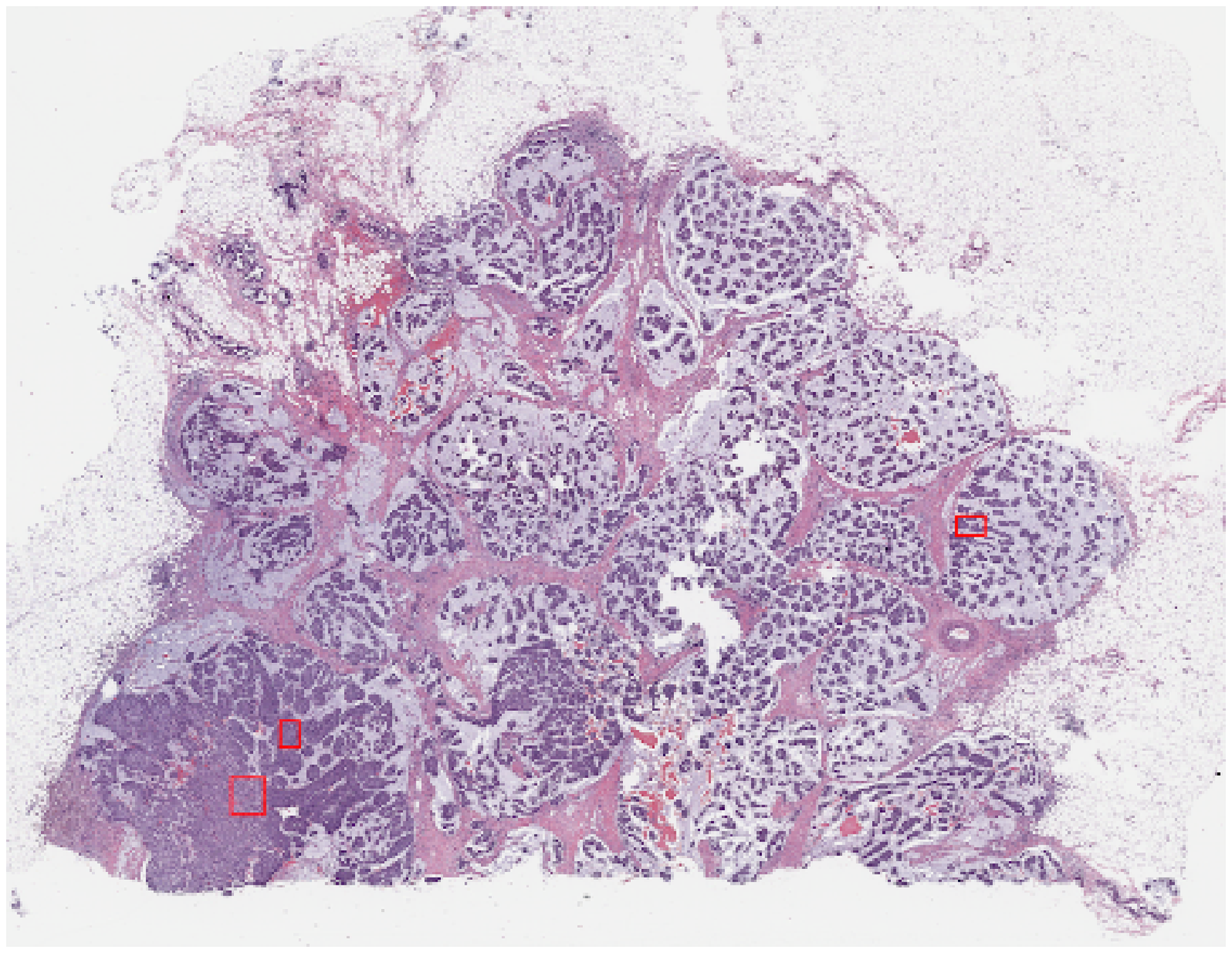}}
      {\includegraphics[width=186pt]{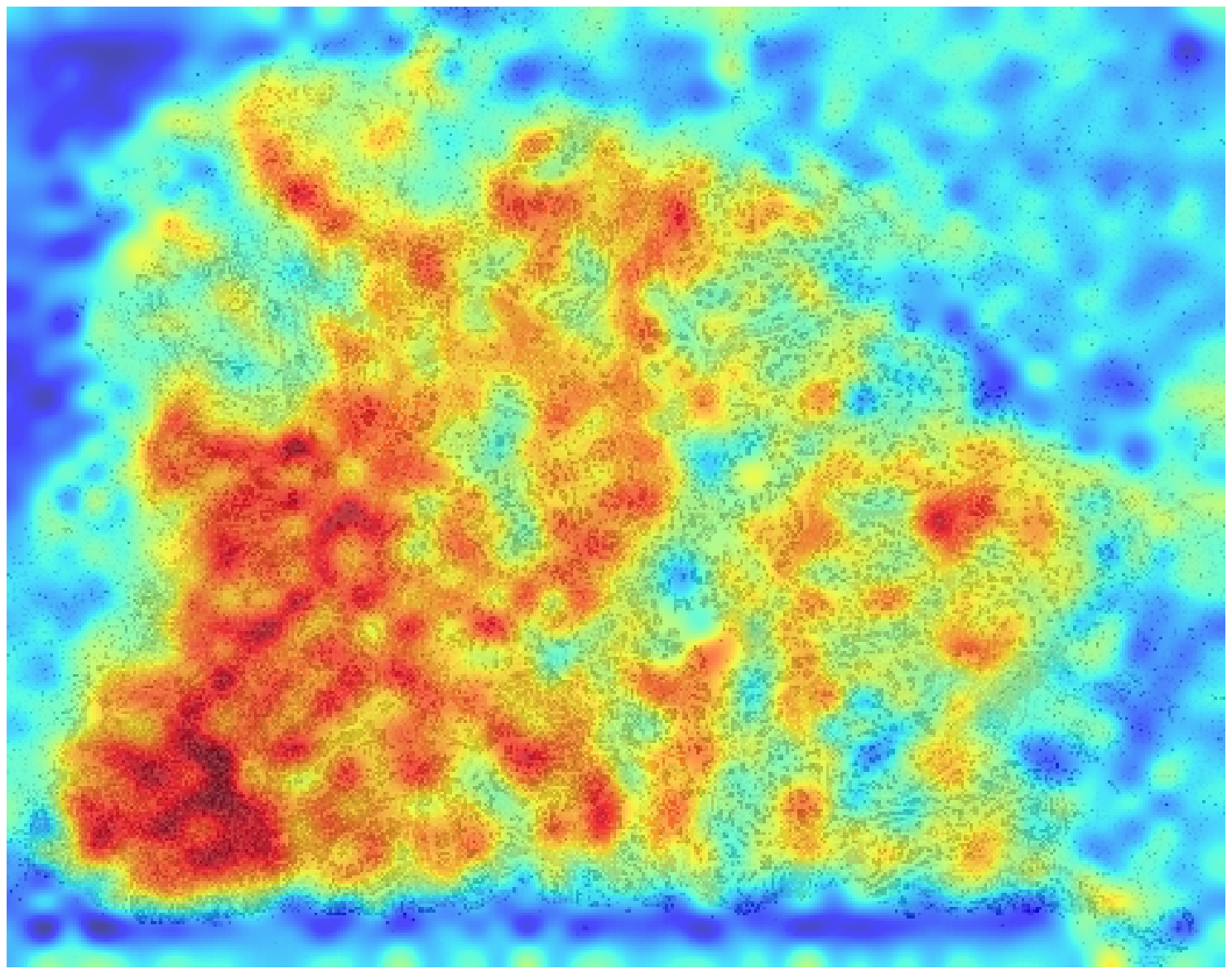}}\\
     \vspace{2cm}

      {\includegraphics[width=190pt]{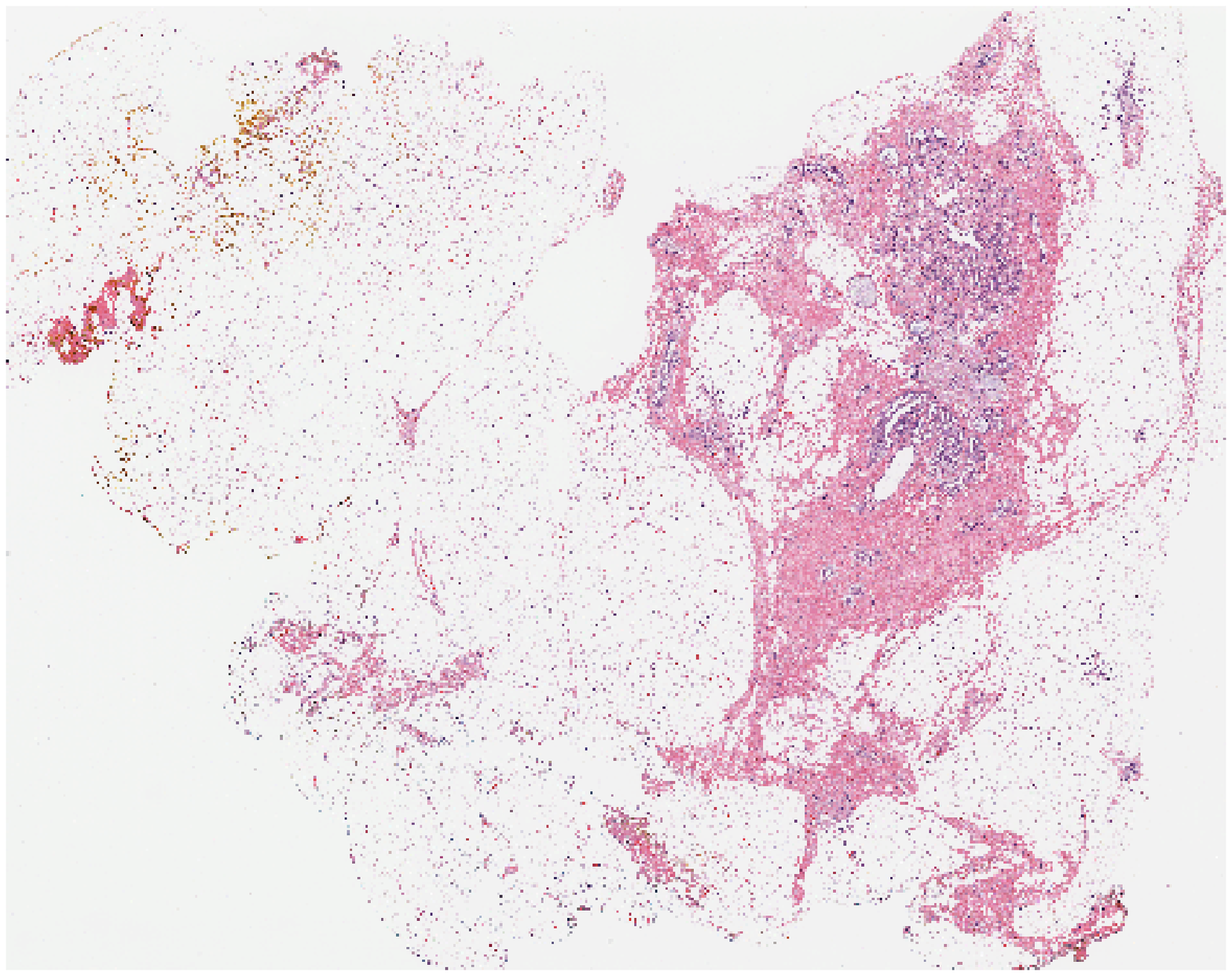}}
      {\includegraphics[width=190pt]{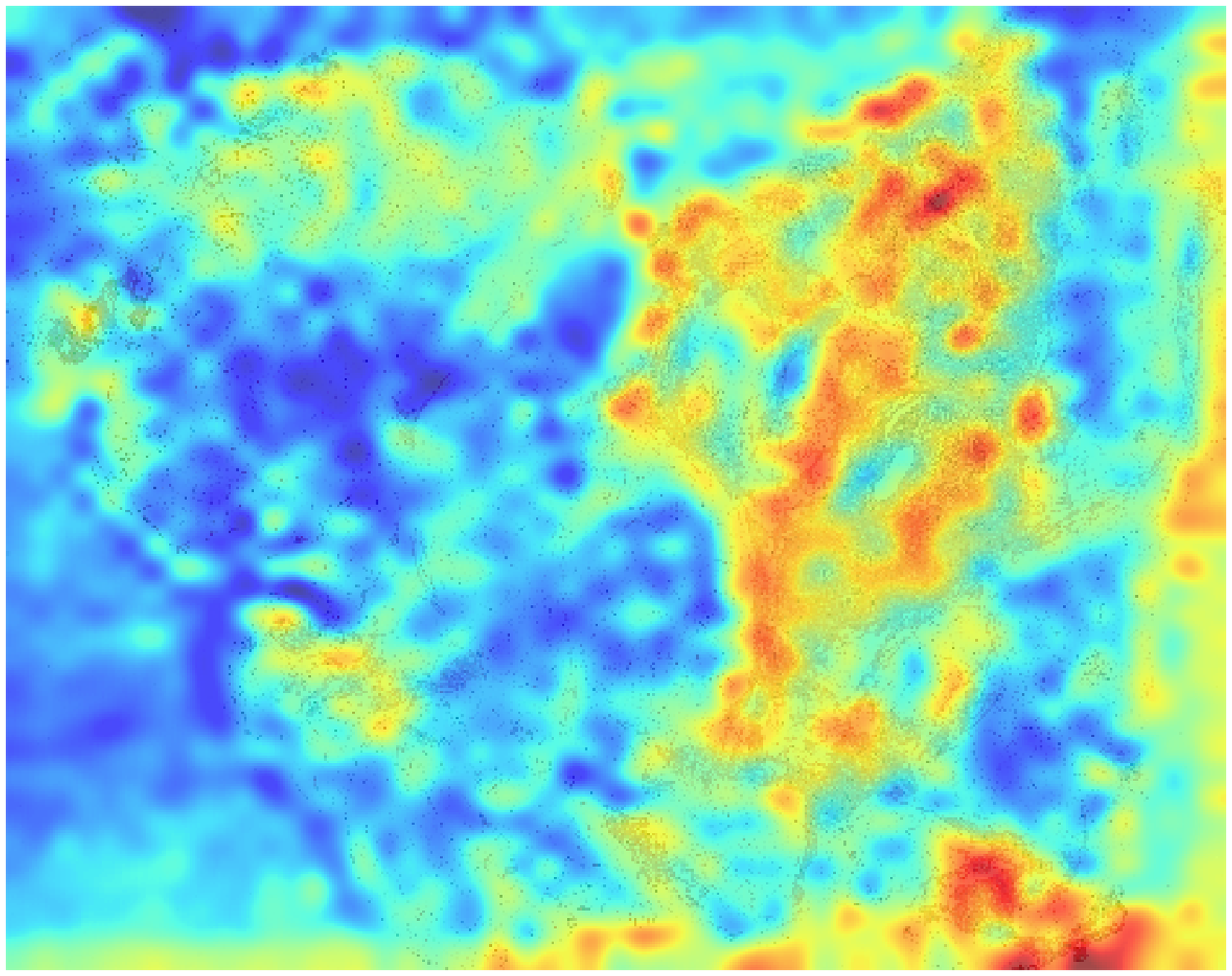}}\\

      \caption{Grad-CAM visualization applied to a sample WSIs from TUPAC16. Dark blue coloring indicates a low saliency, whereas red indicates a high saliency.}
     \label{figure:Saliency2}
\end{figure*}

Both methods ~\cite{tellez2019neural} and ~\cite{tomita2019attention} generate a grid-based compressed version $G$ of the input WSI, even if they adopt different networks and training strategies for the patch feature extraction. Several unsupervised strategies were analyzed in~\cite{tellez2019neural} to achieve such a representation, but the strategy based on the training of a bidirectional generative adversarial network was superior to  the other networks across all experiments with histopathological data. 
In ~\cite{tomita2019attention} the compressed version $G$ is based on features produced by a ResNet18 that is included in the whole training loop of the model.
Our strategy to compute $G$ is based on a simple pre-trained ResNet34 and differently from~\cite{tomita2019attention} it is not included in the whole training loop. On the other hand, our classification model is more complex than that proposed in~\cite{tellez2019neural} that consists of a standard convolutional neural network trained to predict image-level labels. However, the CNN adopted by Tellez et al. is not able to process input of different size, so that compressed images of fixed size are generate by means of crop mechanisms. Our classification model does not undergo this limitation, as it is able to process WSIs of any size. Our architecture incorporates two separate sets of attention maps for a more effective feature learning, while in~\cite{tomita2019attention} only one attention map is adopted. The integration of two sets better helps the network to focus on critical image regions, as well as to highlight discriminative feature channels while suppressing the irrelevant information with respect to the actual classification task.  
Even though there are methods exploring the combination of different attention-based mechanisms \cite{woo2018cbam,huang2019evidence,zhang2019classifying}, they generally consider max- and average-attention maps. However, we believe that max- and average-attention maps are closely correlated. Furthermore, average-attention maps are quite less informative, as they condensate all features in a single value. On the contrary, max- and min-attention maps performs such a kind of feature selection by awarding the most (or least) significant.
In particular, as demonstrated in \cite{courtiol2018classification, durand2017wildcat}, min-attention maps focus on boundary patches between the critical regions highlighted by the max-attention maps, so providing a better discretization.
Indeed, the two different sets of attention maps compute complementary information and are inserted into the network collaboratively.
The attention mechanism adopted in our approach allows to achieve better results than~\cite{tellez2019neural} for both tasks, even if this latter implements a more complex network to generate the compressed image and requires a massive dataset augmentation strategy. Moreover, even if the method of Tomita et al. includes the feature extraction network in the whole training loop of the model, with our approach we obtain better results for task 1 and a comparable performance for task 2. 
We also tested different attention mechanisms and we assessed them both individually and in combination. In particular, in Table~\ref{table:table-pooling}, AUC values are reported when adopting only one Max-pooling (FT-R34-AC Max) or Average pooling (FT-R34-AC Avg) attention map as well as the combination of these attention maps (FT-R34-AC Avg-Max).
As regards the fine-tuning, we observed that the fine-tuned Resnet34 performed better than the simple pre-trained one on Camelyon16, but it provides a lower performance on the TUPAC16 dataset. Accordingly, we have considered that our fine-tuning was performed with respect to the same binary classification task on the Camelyon16 dataset. On the contrary, the task on TUPAC16 is the prediction of the proliferation score and involves only tumoral images. 

\begin{table} [!hb]
\centering
\resizebox{0.4\textwidth}{!}{
	\begin{tabular}{ | c | c |} 
	
	\hline   \textbf{Method} & \textbf{AUC}  \\ 
	\hline FT-R34-AC Max& $0.698 (0.001)$ \\ 
	\hline FT-R34-AC Avg& $0.639 (0.001)$ \\
	\hline FT-R34-AC Avg-Max& $0.654 (0.001)$ \\
	\hline
	\end{tabular}}
	\caption{Comparisons of different Attention map mechanisms.}
	\label{table:table-pooling}
\end{table}
According to the Gradient-weighted Class Activation Mapping (Grad-CAM) method proposed in paper~\cite{selvaraju2017grad}, the spatial position of visual cues relevant in predicting the image-level labels of some WSI samples are shown in Figures~\ref{figure:Saliency1} and ~\ref{figure:Saliency2}. In particular, Figure~\ref{figure:Saliency1} shows comparisons of the saliency maps with fine-grained manual annotations of some examples of WSIs of Camelyon16, while Figure~\ref{figure:Saliency2} reports some examples for TUPAC16. It is worth to notice that in tumor WSIs our model focuses on very specific areas corresponding to active tumor regions.
Additionally, the saliency maps might be exploited by pathologists as a suggestion of regions that need to be analyzed more carefully.

\section{Conclusions}
\label{sec:conclusion}
In this paper, we have proposed a method to analyze gigapixel histopathological images that is trained on weak labeled data. The model compresses the input WSI into a compact representation by rearranging patch-wise feature vectors in a grid-based representation. Two complementary attention based mechanisms are then applied to strengthen discriminant features, while suppressing the noisy ones. A visual inspection of the attention maps confirms that both modules infer discriminant features and lead the network to focus on critical regions correctly, so increasing the accuracy of the tumor evidence localization. 
The performance of the model has been assessed with respect to classification and regression tasks and competitive results have been achieved for both. Thus we believe that such a method shows the potential of a significant impact in diagnosing diseases and predicting lesion positions in histopathological images, saving expert pathologists from time consuming manual annotations. Moreover, it not only improves the state of the art with respect to the classification accuracy, but it also support pathologists by pointing out regions in the WSI that have mainly contributed to take the final decision. Future works will be devoted to assess the potential of this model to scale up with respect to the number of classes in multiclassification tasks.


\end{document}